\pdfoutput=1

\documentclass[11pt]{article}

\usepackage{ACL2023}

\usepackage{times}
\usepackage{latexsym}

\usepackage[T1]{fontenc}

\usepackage[utf8]{inputenc}
\usepackage{enumitem}
\usepackage{microtype}

\usepackage{multirow}
\usepackage{multicol}

\usepackage{graphicx}
\usepackage{bm}
\usepackage{bbm}
\usepackage{amsmath}
\usepackage{amsfonts}
\usepackage{booktabs}
\usepackage{tikz}

\usepackage{makecell}
\usepackage{float}
\usepackage{subfigure}
\usepackage{color}
\usepackage{amssymb}
\usepackage{utfsym}

\newcommand{\modi}{\textcolor{black}}

\title{Outdated Issue Aware Decoding for Reasoning Questions on \\ Edited Knowledge}

\author{
  Zengkui Sun\textsuperscript{1}\thanks{ \ \ Work was done when Zengkui Sun was an intern at Pattern Recognition Center, WeChat AI, Tencent Inc, China.},
  Yijin Liu\textsuperscript{2},
  Jiaan Wang\textsuperscript{3},
  Fandong Meng\textsuperscript{2}, \\
  \textbf{Jinan Xu}\textsuperscript{1},
  \textbf{Yufeng Chen}\textsuperscript{1}\thanks{ \ \ Yufeng Chen is the corresponding author.}
  and \textbf{Jie Zhou}\textsuperscript{2} \\
 \textsuperscript{1}Beijing Jiaotong University, China \\
  \textsuperscript{2}Pattern Recognition Center, WeChat AI, Tencent Inc, China \\
  \textsuperscript{3}Soochow University \\
  \small\texttt{\{zengksun,jaxu,chenyf\}@bjtu.edu.cn} \quad \small{\texttt{jawang.nlp@gmail.com}} \\
  \small{\texttt{\{yijinliu,fandongmeng,withtomzhou\}@tencent.com}} \\
}

\begin{document}
\maketitle
\begin{abstract}
Recently, Knowledge Editing has received increasing attention, since it could update the specific knowledge from outdated ones in pretrained models without re-training.
However, as pointed out by recent studies, existing related methods tend to merely memorize the superficial word composition of the edited knowledge, rather than truly learning and absorbing it.
Consequently, on the reasoning questions, we discover that existing methods struggle to utilize the edited knowledge to reason the new answer, and tend to retain outdated responses, which are generated by the original models utilizing original knowledge.
Nevertheless, the outdated responses are unexpected for the correct answers to reasoning questions, which we named as the outdated issue. 
To alleviate this issue, in this paper, we propose a simple yet effective decoding strategy, \emph{i.e.}, out\textbf{D}ated \textbf{IS}sue aware de\textbf{CO}ding (\textbf{DISCO}), to enhance the performance of edited models on reasoning questions. 
Specifically, we capture the difference in the probability distribution between the original and edited models.
Further, we amplify the difference of the token prediction in the edited model to alleviate the outdated issue, and thus enhance the model performance w.r.t the edited knowledge. 
%
Experimental results suggest that applying DISCO could enhance edited models to reason, \emph{e.g.}, on reasoning questions, DISCO outperforms the prior SOTA method by 12.99 F1 scores, and reduces the ratio of the outdated issue to 5.78\% on the zsRE dataset.
\end{abstract}

\section{Introduction}
Large Language Models (LLMs) exhibit the remarkable ability to capture plenty of factual knowledge into their parameters.
However, knowledge of inner LLMs may become outdated or unsuitable over time~\citep{zhao2021calibrate, elazar2021measuring,dhingra2022time}.
%
Unfortunately, naively re-training LLMs can be computationally intensive, and fine-tuning LLMs in several cases suffers the risk of catastrophic forgetting~\citep{parisi2019continual, mitchell2021fast, ramasesh2021effect}.
Recently, Knowledge Editing has received increasing attention, which aims to update factual knowledge into LLMs and retain the unrelated knowledge, without retraining from scratch~\citep{de2021editing, mitchell2021fast, yao2023editing, mazzia2023survey, yin2023history, zhang2024comprehensive}. 
\begin{figure}[t!]
\begin{center}
    \resizebox{0.475\textwidth}{!}{
        \includegraphics[width=1\textwidth]{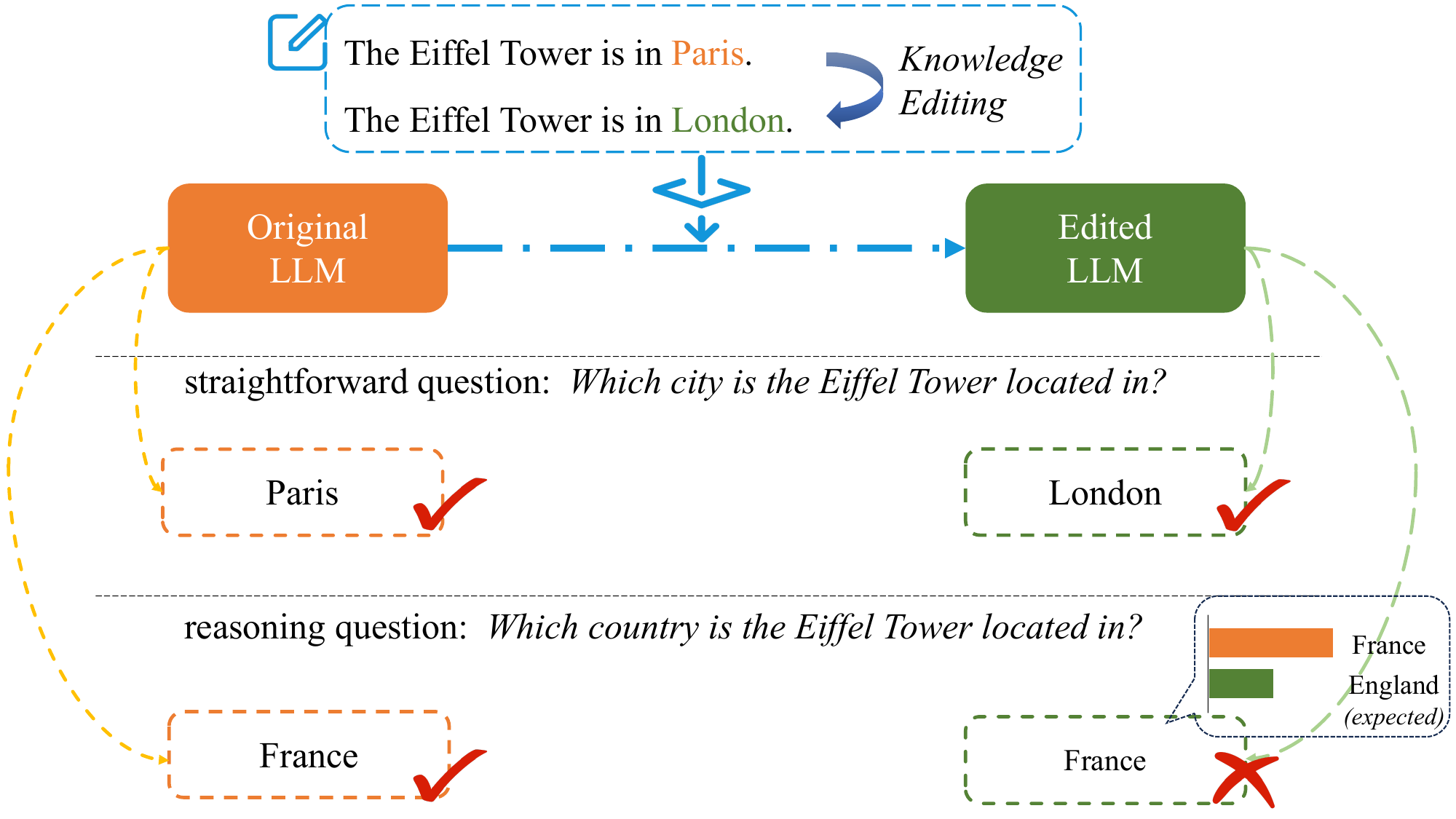}
    } 
    \caption{
        Illustration of the outdated issue in the edited model on reasoning question.
        After knowledge editing, the edited LLM should respond \textit{England} (Country of London) rather than \textit{France} (Country of Paris), where \textit{France} is an outdated response. } 
    \label{fig: showcase}  
\end{center} 
\end{figure}

Previous methods could be divided into two research lines to implement the knowledge editing in LLMs, according to whether preserving original models' parameters~\citep{yao2023editing, mazzia2023survey, wang2023knowledge}.
The first research line retains the weights of the original LLMs, and appends supplementary weights or memories to retrieve the relative edited knowledge to guide the model to respond to the edited knowledge.
For instance, SERAC \citep{mitchell2022memory} uses a scope classifier to determine whether the current prompt falls within the scope of any stored knowledge, and T-Patcher \citep{huang2023transformer} adds extra trainable parameters into FFN layers of LLMs to edit model performance.
Besides, MemPrompt \citep{madaan2022memory}, IKE \citep{zheng2023can}, and MeLLo \citep{zhong2023mquake} conduct in-context learning with knowledge editing demonstrations to prompt LLMs to update pretrained knowledge.
The second research line focuses on adjusting knowledge-related weights in the original LLMs via pre-computation or an additional module to predict the weights.
For instance, KE \citep{de2021editing} trains a BiLSTM to predict the weight update, to enable constrained optimization to modify facts without affecting other knowledge, while KN \citep{dai2021knowledge}, ROME \citep{meng2022locating}, and PMET \citep{li2023pmet} implement editing by first locating the parameters, which store corresponding knowledge, and then directly updating them.
Overall, with these techniques, knowledge editing could update LLMs on new factual edited knowledge without explicit model re-training, when questions explicitly mention the edited knowledge.

However, for reasoning questions, most of these approaches struggle to generate the correct response, and tend to retain the outdated response.
As the example shown in Figure~\ref{fig: showcase}, if we update the knowledge from ``\textit{The Eiffel Tower is in \textbf{Paris}}'' to ``\textit{The Eiffel Tower is in \textbf{London}}'', 
and then query the edited model with the reasoning question ``\textit{Which \textbf{country} is the Eiffel Tower located in?}''.
We find that the edited model still tends to respond with the outdated answer ``\textit{\textbf{France}}'' (Country of ``Paris'') 
rather than the new correct answer ``\textit{\textbf{England}}'' (Country of ``London''), which we named as the outdated issue.
As pointed out by recent studies~\citep{yao2023editing, zhong2023mquake, wang2023cross}, existing methods may tend to memorize its superficial word composition, rather than learning the edited knowledge indeed, \emph{i.e.}, hard coding them into the model locally.

To mitigate the outdated issue, in this paper, we make efforts to explore whether edited models tend to generate outdated responses to reasoning questions.
By analysis, we discover that the edited knowledge exerts a constrained impact on the probability distribution of predicted tokens.
In other words, models tend to utilize their original knowledge to respond to the reasoning questions after editing, which hinders the performance on reasoning questions.
Hence, we propose a decoding strategy, out\textbf{D}ated \textbf{IS}sue aware de\textbf{CO}ding (\textbf{DISCO}), to amplify the impact of edited knowledge on the probability distribution.
Specifically, we capture the modification of probability distribution, by the subtraction of distribution between the original and edited models.
Subsequently, we add this modification to the edited model's probability distribution. 
Besides, we add a constraint to revise the modification of distribution to avoid the probability increase in outdated responses.
In this way, we could amplify the impact of the edited knowledge on the edited model, and alleviate the outdated issue, thus encouraging the edited model to utilize the edited knowledge to reason the new answer.

We conduct experiments to evaluate DISCO on the zsRE dataset~\citep{levy2017zero, yao2023editing} and the CounterFact dataset~\citep{meng2022locating, yao2023editing}.
Experimental results demonstrate that DISCO could effectively mitigate the outdated issue, and then enhance the performance of edited models on reasoning questions.
%
For instance, compared to the prior SOTA method IKE~\cite{zheng2023can}, DISCO improves 12.99 F1 scores and reduces the rate of outdated tokens from 8.23\% to 5.78\% in the zsRE dataset using the \texttt{LlaMa-2-7b} backbone. 
On the other dataset or backbone, DISCO yields the best performance on reasoning questions and suffers the least outdated issue, demonstrating the effectiveness of DISCO.

The main contributions of this paper can be summarized as follows\footnote{Codes are released at \url{https://github.com/Acerkoo/DISCO}.}:
\begin{itemize}[leftmargin=*,topsep=0pt]
\setlength{\itemsep}{0pt}
\setlength{\parsep}{0pt}
\setlength{\parskip}{0pt}
\item To our knowledge, we are the first to point out the edited models suffer from the outdated issue and quantize this issue, which is up to 50\% error ratio among current methods.
\modi{
\item We propose a simple yet effective strategy, \emph{i.e.}, DISCO, to enhance the performance of edited models on reasoning questions. This strategy could amplify the influence of the edited knowledge, and alleviate the outdated issue.
}
%
\item Experimental results show that DISCO could effectively mitigate the outdated issue, and enhance the performance in reasoning questions, without updating parameters.
%
\end{itemize}

\section{Task Formulation} \label{sec: orientation}
The formal definition of knowledge editing is to update factual knowledge into model parameters, which could motivate the model's behavior towards the edited knowledge.

To implement knowledge editing, current researchers mainly follow the two lines:
(1) Retaining the weights of original LLMs, and supplying additional weights or memories.
For instance, SERAC~\citep{mitchell2022memory} uses a scope classifier to determine whether the current prompt falls within the scope of any edited knowledge, and T-Patcher~\citep{huang2023transformer} adds extra trainable parameters into the FFN layers of LLMs to edit knowledge.
Besides, MemPrompt~\citep{madaan2022memory}, IKE~\citep{zheng2023can}, MeLLo~\citep{zhong2023mquake}, and DeepEdit~\citep{Wang2024DeepEditKE} conduct in-context learning or chain-of-thought with knowledge editing examples to prompt LLMs to update pretrained knowledge.
(2) Adjusting knowledge-related weights in original LLMs.
For instance, KE~\citep{de2021editing} trains a Bi-LSTM to predict and update weight, enabling constrained optimization to modify facts without affecting other knowledge, while KN~\citep{dai2021knowledge}, ROME~\citep{meng2022locating}, and PMET~\citep{li2023pmet} implement editing by locating and directly updating the parameters which store corresponding knowledge.

Given an original pretrained language model $\theta$ and an input-output pair of edited knowledge $(x_e, y_e)$, knowledge editing could create an edited model $\hat{\theta}$, which satisfies the following assessment:
\begin{equation}
    \hat{\theta}(x) = \left\{
    \begin{array}{cc}
    y_e, & x \in \mathcal{X}_e, \\
    \theta (x), & x \notin \mathcal{X}_e,
    \end{array}
    \right.
\end{equation}
    
%
where $\mathcal{X}_e$ is the editing scope, denoting a broad set of inputs closely associated with the edited knowledge pair, with similar semantics as $x_e$, 
$\theta(x)$ and $\hat{\theta}(x)$ represent the output of the original model and edited model when receiving the input $x$, respectively. 
Following~\citet{yao2023editing}, the edited model should satisfy the following four properties:
(1) \textbf{Reliability} and (2) \textbf{Generality} straightforward assess the averaged accuracy of the edited case. 
The output of edited model $\hat{\theta}(x)$ should be equal to $y_e$, when the input $x$ is in editing scope $\mathcal{X}_e$.
Note that the difference between Reliability and Generality is that the input for Generality is the paraphrased $x_e$, whereas Reliability is the original $x_e$.
(3) \textbf{Locality} measures the capability of the edited model $\hat{\theta}$ to preserve the performance out of the editing scope. 
That is, $\hat{\theta}(x)$ should be the same as $\theta (x)$ ideally, when $x$ is out of the editing scope $\mathcal{X}_e$.
(4) \textbf{Portability} evaluates the effectiveness of the edited model in transferring edited knowledge to related content.
When receiving an input of reasoning question $x$, which requires reasoning based on the edited knowledge, the edited model is expected to correct answer it to demonstrate the model learns the knowledge itself rather than only memorizing superficial changes in wording.

\section{Findings of the Outdated Issue}
\subsection{Oudated Issue} \label{sec: kl}
Among the above four properties, \textbf{Portability} is more challenging in assessing the effect of knowledge editing, since it requires edited models to learn the edited knowledge indeed and reason on it~\citep{yao2023editing, zhong2023mquake, wang2023cross, ma2023untying, zhang2024comprehensive}.
As reported by prior studies~\citep{yao2023editing, wang2023cross, wang2023easyedit}, ROME~\citep{meng2022locating}, MEMIT~\citep{meng2022mass}, and IKE~\citep{zheng2023can} perform better in Portability.
IKE is in the first research line, utilizing the robust capabilities of LLMs for in-context learning to edit LLMs by prompting with retrieved demonstrations from the external memory. 
ROME and MEMIT are in the second research line, specifying the FFN matrix as the key-value neurons embodying knowledge, and they implement editing by locating knowledge-related neurons and updating them.

Unfortunately, in our preliminary experiments, we discover that the above approaches suffer the \textbf{Outdated Issue} that the edited model $\hat{\theta}$ tends to generate the outdated output $\theta(x)$, in Portability. 
For instance, we evaluate these approaches with\texttt{GPT-J-6b} on the zsRE dataset, and observe that these approaches seriously suffer the issue, ranging from 12\% to 50\% ratio of this issue (please refer to Section~\ref{sec: main_results} for more details).
%

\subsection{Similarity of Probability Distribution} \label{sec: similarity}
Since decoding is directly applied in the response generation stage, to take a further step to probe the outdated issue, we explore the similarity of probability distribution of predicted tokens between the original model $\theta$ and the edited model $\hat{\theta}$.

Given an input $x$, we could model the conditional probability distribution of the $t$-th token: 
\begin{equation}
    P_{\theta}^{t}(\cdot) = p_\theta^{t}(\cdot | x, y_{<t}),
\end{equation}
where $p_{\theta}^{t}(\cdot)$ denotes the conditional probability distribution of the original model $\theta$ for the $t$-th predictied token, $y_{<t}$ denotes the previously predicted tokens by the original model.

Similarly, the probability distribution of the edited model $\hat{\theta}$ could be calculated as follows:
\begin{equation}
    P_{\hat{\theta}}^{t}(\cdot) = p_{\hat{\theta}}^{t}(\cdot | x, \hat{y}_{<t}),
\end{equation}
where $p_{\hat{\theta}}^{t}(\cdot)$ and $\hat{y}_{<t}$ denote the edited model $\hat{\theta}$’s corresponding items of original model $\theta$.

To invest the similarity between $P_{\theta}^{t}(\cdot) $ and $P_{\hat{\theta}}^{t}(\cdot)$, we apply Jensen-Shannon divergence to calculate the distance of both probability distributions, following~\citet{chuang2023dola}: 
\begin{equation}
\begin{aligned}
    \mathrm{JSD} (P_{\theta}^{t}(\cdot), P_{\hat{\theta}}^{t}(\cdot)) = \frac{1}{2}\cdot KL(P_{\theta}^{t}(\cdot) || P_{\hat{\theta}}^{t}(\cdot)) \\ + \frac{1}{2}\cdot KL(P_{\hat{\theta}}^{t}(\cdot) || P_{\theta}^{t}(\cdot)), \quad \quad
\end{aligned}
\end{equation}
where $\mathrm{JSD}(\cdot)$ denotes the function of Jensen-Shannon divergence,
$KL(\cdot)$ denotes the function of Kullback-Leibler divergence.
In this case, the smaller value of $\mathrm{JSD}(\cdot)$ denotes the more similar of both probabilities, suggesting that the less impact of edited knowledge $(x_e, y_e)$ to edited model $\hat{\theta}$'s output distribution.
As a result, the edited model $\hat{\theta}$ will tend to generate the outdated response $\theta(x)$.

\begin{table}[t!]
\centering
\resizebox{0.45\textwidth}{!} {
\begin{tabular}{c|c|ccc}
    \toprule
    \textbf{Property} & \textbf{Method} & \textbf{JSD} & \textbf{OE}$\downarrow$ & \textbf{F1 / EM}$\uparrow$ \\
    \midrule
    Reliability & IKE & 41.62 & - & 99.90 / 99.71 \\
    \midrule
    \multirow{4}*{Portability} & FT & \;\;0.23 & 48.83 & 30.39 / 0.00\;\; \\
    ~ & ROME & 16.83 & 15.62 & 31.70 / 2.70\;\; \\
    ~ & MEMIT & \;\;9.16 & 30.35 & 33.01 / 1.35\;\; \\
    ~ & IKE & 19.15 & 12.41 & 42.05 / 11.39 \\
    \bottomrule
\end{tabular}
}
\caption{
\modi{
    Performance of edited models in Reliability and Portability on the zsRE dataset with the \texttt{GPT-J-6b}.
    %
    JSD denotes the value of Jensen-Shannon divergence, and OE reports the ratio of outdated issues. 
    F1/EM reports the F1 and EM scores in Portability.
}
}
\label{tab: similarity_results}
\vspace{-4pt}
\end{table}

To understand this similarity of the probability distribution better, we analyze the similarity of both models in Reliability and Portability.
As shown in Table~\ref{tab: similarity_results}, compared to the similarity of both models of IKE in Reliability, all edited models' probability distribution have a smaller JSD value and are more similar to the original model in Portability.
Besides, the more similar probability distribution between the original and edited models meets the more serious outdated issue and the worse performance on reasoning questions. 
These results suggest that the edited knowledge takes a few modifications on probability distribution, and the generation of new answers is disturbed by the original knowledge.
Therefore, we should amplify the impact of the edited knowledge on probability distribution, to reduce the disturbance of original knowledge, thus encouraging the edited model to utilize the edited knowledge to reason the new answer.

\section{DISCO: Outdated Issue Aware Decoding}
As analyzed in the prior section (\S\ref{sec: similarity}), the few impacts of edited knowledge on probability distribution and the disturbance of original knowledge during reasoning should be responsible for the outdated issue.
To amplify the impact of the edited knowledge on probability distribution, we propose a simple yet effective method, out\textbf{D}ated \textbf{IS}sue aware de\textbf{CO}ding (\textbf{DISCO}).

We first implement knowledge editing based on the prior editing methods, \emph{e.g.}, IKE, which performs best in Portability~\citep{yao2023editing, wang2023cross}.
Given an input $x$ to the original model $\theta$ and edited model $\hat{\theta}$, we capture their difference in probability distribution, caused by edited knowledge, via the subtraction as follows:
\begin{equation}
\begin{split}
    \mathit{\Delta}(t)
    &= P_{\hat{\theta}}^{t}(\cdot) - P_{\theta}^{t}(\cdot) \\
    &= p_{\hat{\theta}}^{t}(\cdot | x, \hat{y}_{<t}) - p_{\theta}^{t}(\cdot | x, y_{<t}).
\label{equ: delta}
\end{split}
\end{equation}
Then, we add the difference $\mathit{\Delta}(t)$ to the probability distribution of the edited model to amplify the difference in probability distribution:
\begin{equation}
\begin{split}
    p_{\hat{\theta}}^{t}(\cdot) &= p_{\hat{\theta}}^{t}(\cdot |x, \hat{y}_{<t}) + \alpha \cdot \mathit{\Delta}(t).
\end{split}
\label{equ: decoding}
\end{equation}
where $\alpha > 0$ is a hyperparameter to control the weight of $\mathit{\Delta}(t)$.
By this formula, we form a simple contrastive decoding~\citep{li2022contrastive, shi2023trusting, chuang2023dola} between the original and edited models, 
to capture and amplify the difference of edited knowledge in probability distribution, and then prevent the outdated issue.

\noindent \textbf{Constraints to revise Probability.}
To further constrain the probability of tokens of outdated response not increasing, we limit the maximum of $\mathit{\Delta}(\cdot)$ for the tokens in outdated response $\theta(x)$:
\begin{equation} \label{eq.7}
    \mathit{\Delta}(t) = \left\{
    \begin{array}{ll}
         \mathrm{min}(0, \mathit{\Delta}(t)), & \mathcal{V}_{out}, \\
         \mathit{\Delta}(t), & otherwise,
    \end{array} 
    \right.
\end{equation}
where $\mathcal{V}_{out}$ denotes the token set of the outdated response $\theta (x)$.

Furthermore, applying $\mathit{\Delta}(t)$ in Eq.\ref{equ: decoding} faces the risk of increased probability of the target of factual edited knowledge, we take a similar constraint to the tokens of $y_e$:
\begin{equation}
    \mathit{\Delta}(t) = \left\{
    \begin{array}{ll}
         \mathrm{min}(0, \mathit{\Delta}(t)), &  \mathcal{V}_{out} \cup \mathcal{V}_{edit}, \\
         \mathit{\Delta}(t), & otherwise,
    \end{array} 
    \right.
\label{equ: limit}
\end{equation}
where $\mathcal{V}_{edit}$ is the token set of the target of edited knowledge (token-level $y_e$).

\noindent \textbf{Knowledge-aware In-cotext Editing}.
To further enhance the awareness of LLMs to the edited knowledge, we prepend the paraphrase of $x_e$ and the answer $y_e$ to the real input $x$ as an in-context example.
With the example, we could take a further step to amplify the impact of edited knowledge on probability distribution and reduce the disturbance of previous knowledge.

Overall, we design the DISCO strategy, consisting of Eq.\ref{equ: decoding}, Eq.\ref{equ: limit}, and the in-context editing, to amplify the modification in the probability distribution.
In this way, DISCO mitigates the outdated issue, and then encourages the edited model to utilize the edited knowledge to reason the new answer.

\section{Experiments}

\subsection{Metrics} \label{sec: metric}
For Reliability, Generality, and Portability, different types of questions are inputted to the edited model, and we compare the model answers with ground truth ones to calculate token-level F1 and exact match (EM), following previous QA studies~\citep{rajpurkar-etal-2016-squad, yang-etal-2018-hotpotqa}:
(1) F1 measures the average overlap between the prediction and the golden answer;
(2) EM measures the percentage of predictions which exactly match the golden answer.
For Locality, an irrelevant question is used to evaluate whether the edited model retains the original performance.
Hence, the golden answer of Locality is the original output.
We also calculate the token-level F1 and EM to evaluate the edited models in terms of Locality.

Furthermore, when using portability to evaluate the edited models, we quantify the outdated error ($\mathit{OE}$) by calculating the averaged proportion of outdated tokens (each of which appears in the outdated response $\theta (x)$).
To avoid over-count, we remove the overlap tokens between the ground truth and edited model predictions when calculating \textit{OE}:
\begin{equation}
    \mathit{OE}(\hat{\theta}) = \frac{1}{m} \sum_{t=1}^{m}\mathbbm{1} \{\hat{y}_t \in \mathcal{V}_{out} \;\mathrm{\&}\; \hat{y}_t \notin \mathcal{V}_{golden}\},
\label{eq: oe}
\end{equation}
where $m$ is the number of generated tokens, $\mathcal{V}_{out}$ and $\mathcal{V}_{golden}$ denote the token set of outdated response and golden answer in Portability, respectively.

Moreover, we detect the ratio of target tokens of edited knowledge in the predictions, noted as $\textit{TE}$:
\begin{equation}
    \mathit{TE}(\hat{\theta}) = \frac{1}{m} \sum_{t=1}^{m}\mathbbm{1} \{\hat{y}_t \in \mathcal{V}_{edit} \;\mathrm{\&}\; \hat{y}_t \notin \mathcal{V}_{golden}\}.
\end{equation}

\begin{table*}[t!]
\centering
\resizebox{0.95\textwidth}{!} {
    \begin{tabular}{c|c|ccc|c|cc}
        \toprule
        \textbf{Backbone} & \makecell[c]{\textbf{Method}} & \textbf{Reliability}$\uparrow$  & \textbf{Generality}$\uparrow$ & \textbf{Locality}$\uparrow$ & \textbf{Portability}$\uparrow$ & \textbf{OE}$\downarrow$ & \textbf{TE}$\downarrow$ \\
        \bottomrule
        \multicolumn{8}{c}{\textit{zsRE}} \\
        \toprule
        \multirow{3}*{\makecell[c]{\texttt{GPT-J-6b}}} & FT & 15.32 / 0.10\;\; & 14.73 / 0.00\;\; & 99.22 / 97.40 & 30.39 / 0.00 & 48.83 & \;\;\textbf{1.28} \\
        ~ & IKE & \textbf{99.90} / \textbf{99.71} & \textbf{99.82} / \textbf{99.61} & 52.28 / 28.93 & 36.46 / 3.57 & 12.41 & 25.80 \\
        ~ & DISCO & 98.30 / 97.59 & 97.32 / 96.14 & \textbf{54.97} / \textbf{31.44} & \textbf{42.05} / \textbf{11.39} & \textbf{11.58} & \textbf{13.78} \\
        \midrule
        \multirow{3}*{\makecell[c]{\texttt{LlaMa-2-7b}}} & FT & 43.29 / 9.35\;\; & 36.80 / 4.34\;\; & 93.51 / 80.62 & 35.21 / 0.77 & 12.75 & 10.90 \\
        ~ & IKE & \textbf{99.84} / \textbf{99.71} & \textbf{99.63} / \textbf{99.32} & 56.27 / 22.08 & 50.88 / 10.90 & \;\;8.23 & 17.64 \\
        ~ & DISCO & 99.02 / 98.17 & 98.90 / 97.69 & \textbf{62.64} / \textbf{30.95} & \textbf{63.87} / \textbf{33.46} & \;\;\textbf{5.78} & \;\;\textbf{6.09} \\
        \bottomrule
        \multicolumn{8}{c}{\textit{CountFact}} \\
        \toprule
        \multirow{3}*{\makecell[c]{\texttt{GPT-J-6b}}} 
        & FT & \;\;5.04 / 5.04\;\; & \;\;0.97 / 0.97\;\; & \textbf{96.12} / \textbf{96.12} & 25.76 / 0.00 & 55.75 & \;\;\textbf{0.02} \\
        ~ & IKE & \textbf{99.71} / \textbf{99.71} & 74.78 / 74.78 & 20.47 / 20.56 & 28.48 / 0.00 & 49.57 & \;\;0.06 \\
        ~ & DISCO & 90.30 / 90.30 & \textbf{86.52} / \textbf{86.52} & 15.13 / 15.23 & \textbf{29.40} / 0.00 & \textbf{45.88} & \;\;\textbf{0.02} \\
        \midrule
        \multirow{3}*{\makecell[c]{\texttt{LlaMa-2-7b}}}
        & FT & 37.15 / 26.67 & 27.60 / 19.30 & \textbf{38.60} / \textbf{31.23} & 30.21 / 0.00 & 42.92 & \;\;1.65 \\
        ~ & IKE & \textbf{99.44} / \textbf{99.32} & \textbf{82.00} / \textbf{79.44} & 25.86 / 17.75 & 38.05 / 0.00 & 37.49 & \;\;0.33 \\
        ~ & DISCO & 91.85 / 90.40 & 81.43 / 78.56 & 19.13 / 9.80\;\; & \textbf{39.41} / 0.00 & \textbf{36.34} & \;\;\textbf{0.08} \\
        \bottomrule
    \end{tabular}
}
\caption{
    Experimental results (F1/EM) on the \textit{zsRE} and \textit{CounterFact} datasets with \texttt{GPT-J-6b}, \texttt{LlaMa-2-7b}.
    All digital results denote the token-level score of the corresponding property.
    \textbf{Portability} is the core problem in this paper.
    \textbf{DISCO} denotes our decoding strategy.
    \textbf{Bold} denotes the best performance.
}
\label{tab: main_results}
\end{table*}

\subsection{Experimental setup}
\noindent \textbf{Datasets.} 
In our experiments, we mainly conduct experiments on zsRE~\citep{levy2017zero}, which is one of the most prevalent Question Answering datasets extended and adopted for knowledge editing~\citep{de2021editing, mitchell2021fast}.
Further, \citet{yao2023editing} expand the zsRE test set on portability, in terms of \textit{subject replacement}, \textit{inverse relation}, and \textit{one-hop}.
To evaluate the edited models on reasoning questions, we carefully select \textit{one-hop}, which requires models to take one-step reasoning based on the edited knowledge.

Moreover, we also evaluate our methods on CounterFact~\citep{meng2022locating, yao2023editing}, which is a more challenging dataset that consists of counterfactual edits.
For instance, the CounterFact updates the knowledge ``\textit{Apple A5 was created by \textbf{Apple}}'' to ``\textit{Apple A5 was created by \textbf{Google}}''.

\noindent \textbf{Backbones.}
Following prior studies~\citep{meng2022locating, meng2022mass, yao2023editing}, we adopt the \texttt{GPT-J-6b}~\citep{gpt-j}, \texttt{LlaMa-2-7b} and \texttt{LlaMa-2-13b}~\citep{touvron2023llama} as the backbones.

\noindent \textbf{Baselines.}
We adopt four methods as baselines:
(1) directly fine-tuning (\textbf{FT}) the language models with $L_{\infty}$ constraint;
(2) \textbf{ROME}~\citep{meng2022locating} leverages causal mediation analysis to locate the edit area, and updates the whole parameters in the FFN matrix;
(3) \textbf{MEMIT}~\citep{meng2022mass} directly updates LLMs with many memories, thus editing thousands of knowledge simultaneously;
(4) \textbf{IKE}~\citep{zheng2023can} using in-context learning to guide models to update knowledge.

\noindent \textbf{Implementation Details.}
All experiments are conducted on a single NVIDIA A100 GPU (40G).
%
The implementation of all baselines and our method is employed by EasyEdit~\citep{wang2023easyedit}, with the default hyper-parameters in the GitHub repository\footnote{\url{https://github.com/zjunlp/EasyEdit/tree/main/hparams}}. 
Since \modi{in-context editing} methods perform best in Portability~\citep{yao2023editing, wang2023cross}, we implement DISCO via the in-context learning and set $\alpha$ to 1.0\footnote{We explore the impact of $\alpha$ in Appendix.\ref{appendix: hyper}.}.
To conduct in-context learning examples, we search for the most related example of knowledge editing case to guide the edited model.
Similar to IKE~\citep{zheng2023can}, we apply \texttt{all-MiniLM-L6-v2}\footnote{\url{https://huggingface.co/sentence-transformers/all-MiniLM-L6-v2}} to encode and retrieve examples of knowledge editing cases.

\subsection{Main Results} \label{sec: main_results}
We conduct the main experiments on the zsRE and CounterFact datasets, and then report the F1 and EM scores of four types of questions and the ratio of \textit{OE} and \textit{TE} in Table~\ref{tab: main_results}.
We find that IKE has the best performance in Portability, which is consistent with previous studies \citep{yao2023editing, wang2023cross}. 
Besides, IKE and DISCO are in-context editing methods, without any parameters updating.
Hence, we mainly compare DISCO with IKE.
We discuss the performance of ROME and MEMIT in Appendix~\ref{sec: full}.

As shown in Table \ref{tab: main_results}
(1) In terms of Reliability, on both datasets, DISCO has achieved great success in yielding over 90 F1 and EM scores with both backbones.
For the Generality, DISCO yields over 98.90 and 81.43 F1 scores on the zsRE and CounterFact datasets, respectively.
This indicates that our DISCO is competitive with the previous approaches, when the questions are within the editing scope.
Note that, for both IKE and DISCO methods, Reliability and Generality mainly measure the ability to memorize the edited knowledge from the prompt, since the edited knowledge in the prompt has a similar or same format as the input question.
(2) For the Locality, 
both in-context editing methods perform poorly on both datasets.
Although FT could maintain the performance to the irrelevant question, we consider the performance comes from the invalid knowledge editing, where most F1 scores of FT are less than 40 in Reliability.
DISCO performs better than IKE in the zsRE dataset, while worse than IKE in the CounterFact dataset.
We conjecture this poor Locality performance is attributed to that LLMs struggle to locate the editing scope and affect unrelated inputs~\citep{yao2023editing}, and could be alleviated by supplying a scope classifier to determine whether editing model~\citep{mitchell2022memory}.

For the reasoning properties:
(1) In Portability, DISCO yields the best performance among all methods on both backbones.
Compared to IKE, DISCO improves F1 score with +5.59 and +12.99 scores on the \texttt{GPT-J-6b} and \texttt{LlaMa-2-7b}, respectively.
Furthermore, the EM score of DISCO is around triple times (11.39 vs. 3.57 and 33.46 vs. 10.90) that of IKE.
In the challenging CounterFact dataset, DISCO still outperforms IKE by over +1.0 F1 score.
Since the challenge, edited models struggle to respond whole golden answer (EM), and only predict a few tokens of the golden answer.
Even so, these improvements demonstrate that DISCO could powerfully assist LLMs to generate correct answers, by capturing and amplifying the modification of the edited knowledge, on reasoning questions.
(2) DISCO also performs best on the Outdated Issue among all methods in terms of \textit{OE} and \textit{TE}, generating the least outdated tokens after knowledge editing.
In the zsRE dataset, DISCO could yield only a 5.78\% average ratio of the outdated issue on the \texttt{LlaMa-2-7b}.
As the results show, DICSO could effectively prevent the edited model from generating outdated tokens and target tokens of edited knowledge, simultaneously.

Moreover, when comparing edited models on different datasets or backbones, we could find that the basic ability of backbones and the difficulty of editing knowledge are two important factors in the performance of edited models, especially in Portability.
Overall, DISCO is an effective method, performing competitive results in Reliability and Generality, and achieving new state-of-the-art performance in Portability.

\section{Analysis}
\begin{figure}[t!]
\centering
    \resizebox{0.455\textwidth}{!}{
        \includegraphics[width=1\textwidth]{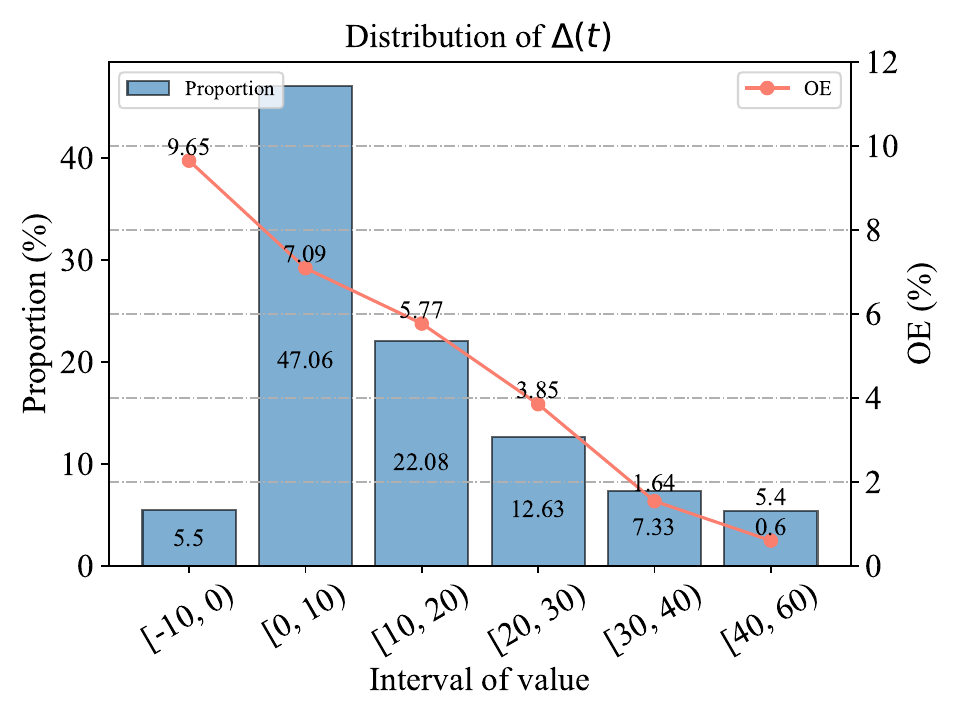}
    } 
\caption{
    \modi{Probability distribution of $\mathit{\Delta}(t)$ in Portability on the zsRE dataset with \texttt{LlaMa-2-7b}.
    Interval denotes the interval of the averaged value of $\mathit{\Delta}(t)$ (\%).}
}
\label{tab: dist_results}
\end{figure}

\subsection{Distribution of $\mathit{\Delta}(t)$} \label{sec: dist}
\modi{We investigate the probability variation $\mathit{\Delta}(t)$ (in Eq.\ref{equ: delta})
to further probe the impact of $\mathit{\Delta}(t)$
on the outdated issue.
On the zsRE dataset and \texttt{LlaMa-2-7b} backbone, 
we split the averaged $\mathit{\Delta}(t)$ value in Portability into intervals with the 10.0\%, 
and then count the proportion and the \textit{OE} metric w.r.t each interval.
}

\modi{
Since the positive or negative $\mathit{\Delta}(t)$ value denotes whether the probability of golden answers is improved, as illustrated in Figure~\ref{tab: dist_results}, we could observe that most value of $\mathit{\Delta}(t)$ of golden answers (around 94.5\%) has a positive value.
The positive value indicates that $\mathit{\Delta}(t)$ could assist the edited model in reasoning the golden answer.
Besides, with the larger $\mathit{\Delta}(t)$ value, the ratio of outdated issue of the corresponding interval has a monotone decreasing trend (from 9.65\% to 0.6\%).
This trend further demonstrates that $\mathit{\Delta}(t)$ could assist the edited model in reasoning the golden answer, and mitigate the outdated issue.
}
\begin{table}[t]
\centering
\resizebox{0.365\textwidth}{!} {
\begin{tabular}{c|c|cc} 
    \toprule
    \multirow{2}{*}{\textbf{Model}} & \multirow{2}{*}{\textbf{Locality}} & \multicolumn{2}{c}{\textbf{Portability}} \\
    ~ & ~ & Golden$\uparrow$ & Outdated$\downarrow$ \\
    \midrule 
    \textbf{Edited} &\textbf{58.03} & 66.44 & 54.14 \\
    \midrule
    \textbf{DISCO} & 56.48 & \textbf{77.33} & \textbf{30.07} \\
    \bottomrule
\end{tabular}
}
\caption{
    \modi{
    Averaged probability of golden answers in Locality and Portability, on the zsRE dataset with \texttt{LlaMa-2-7b}.
    Golden and Outdated denote the value of golden answers and outdated responses, respectively.
    Edited and DISCO represent the performance of the edited model with or without DISCO, respectively.
    }
}
\label{tab: property_results}
\end{table}
\subsection{Variation of Probability Distribution} \label{sec: variation}
\modi{
To probe the variation of probability distribution after applying DISCO, we count the average probability of golden answers in four properties, and the outdated responses in Portability.
We display the results on the zsRE dataset with \texttt{LlaMa-2-7b}, and list the results in Table~\ref{tab: property_results}.
}

\modi{
For the Locality, DISCO suffers a little sacrifice in probability to retain the output of the original model, and this sacrifice could be supplied by a scope classifier or parameter locating.
For Portability, DISCO could enlarge the difference of probability between the outdated response and golden answers from 12.30\% to 47.26\%.
Consequently, DISCO could significantly capture and amplify the impact of edited knowledge on the probability distribution, then alleviate the outdated issue and encourage the edited model to generate the correct answers for reasoning questions.
}

\begin{figure}[t!]
\begin{center}
    \scalebox{0.455}{
        \subfigure[] {
            \begin{minipage}[t]{\linewidth}
            \includegraphics[width=\textwidth]{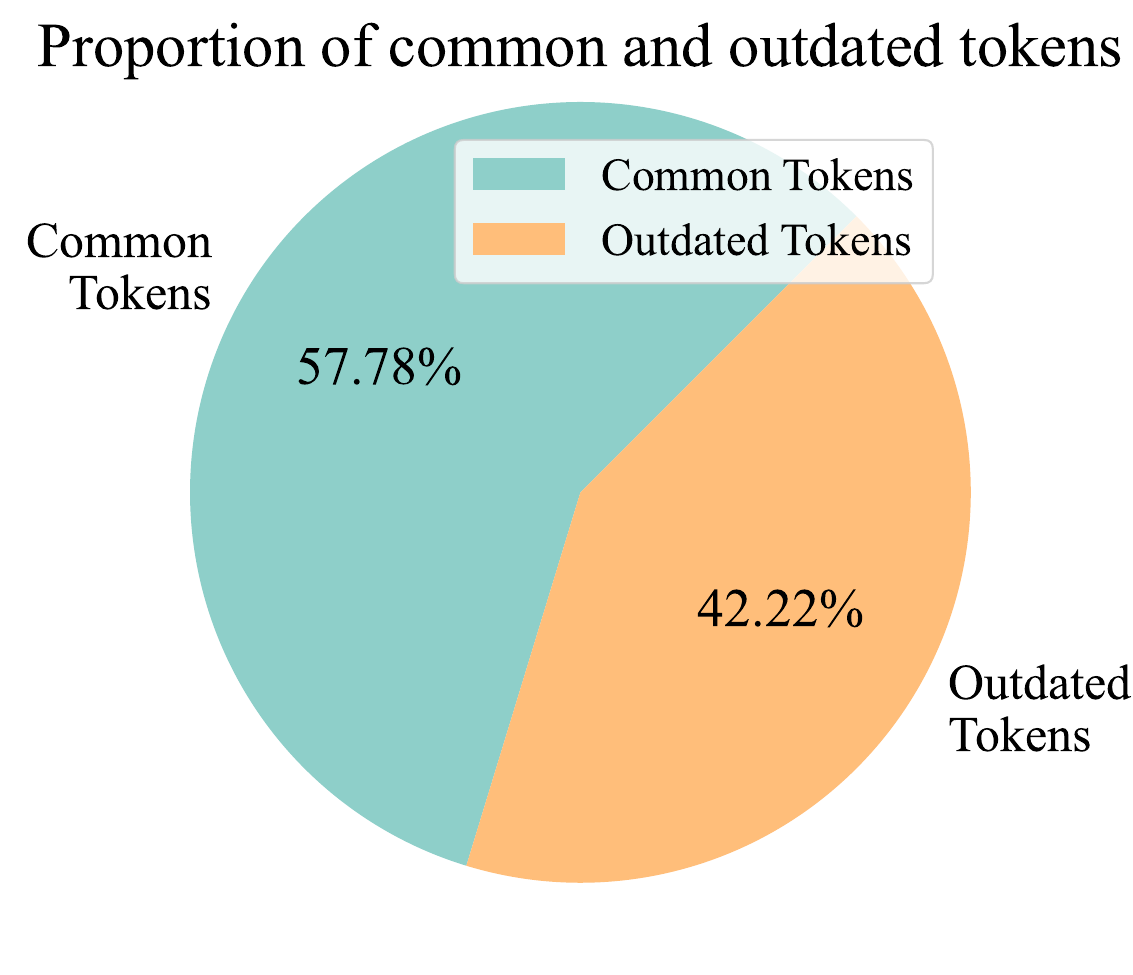}
            \end{minipage}
            \label{fig: tokens_prop}
        }
    \quad
        \subfigure[] {
            \begin{minipage}[t]{\linewidth}
            \includegraphics[width=\textwidth]{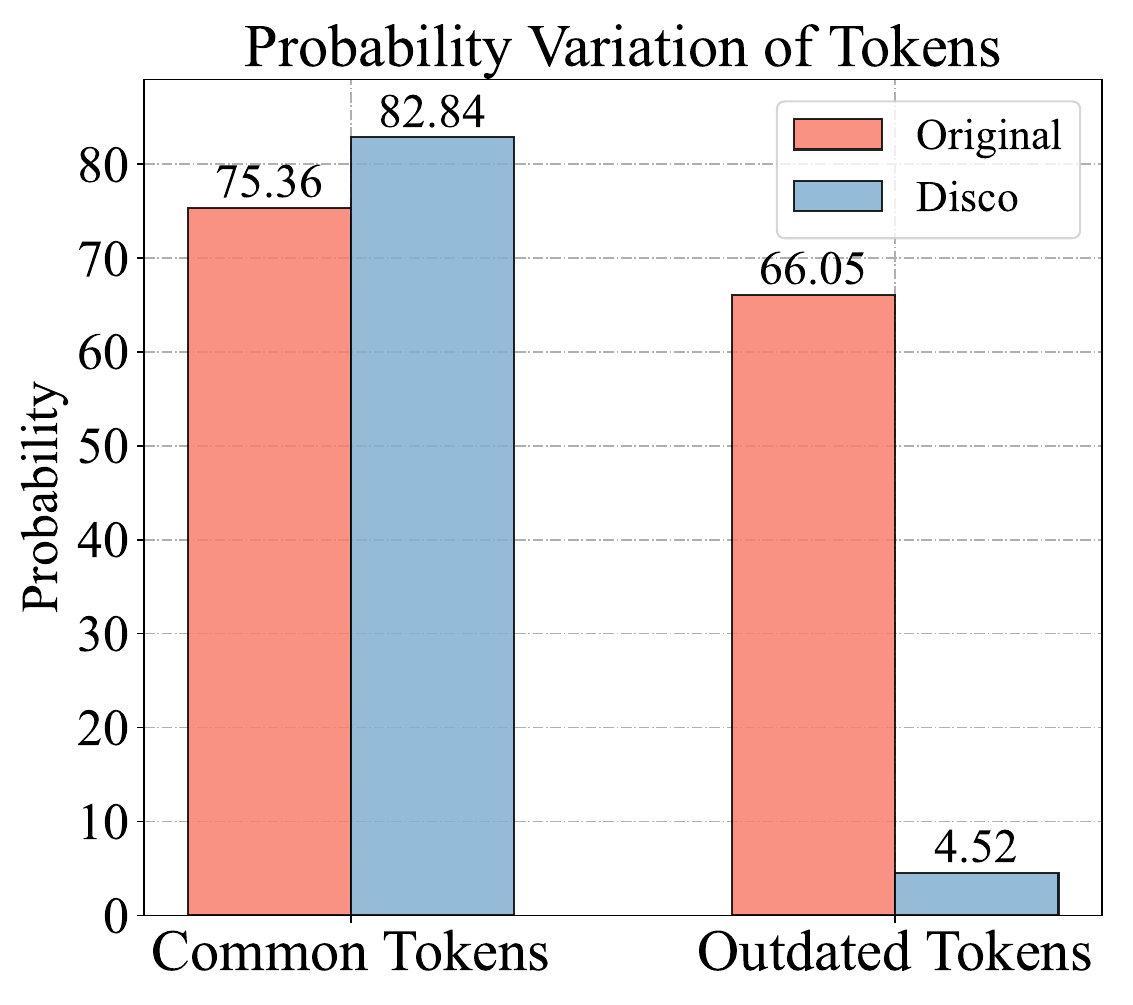}
            \end{minipage}
            \label{fig: tokens_prob}
        }
    }
    \caption{
        \modi{The proportion of common tokens in outdated responses, and the probability variation of the tokens before and after applying DISCO.}
    } 
    \label{fig: tokens_dist}  
    \vspace{-4pt}
 \end{center} 
\end{figure}
\subsection{Probability of Common Tokens} \label{sec: common}
To detect whether all tokens in outdated responses are wrong, on reasoning questions, we count the proportion of the common tokens in both the outdated response and the new golden answer, and their probability variation.
We display the statistical results of DISCO on the zsRE dataset with \texttt{LlaMa-2-7b} in Figure~\ref{fig: tokens_dist}.

As shown in Figure~\ref{fig: tokens_prop}, in the token-level, 57.78\% tokens in outdated responses will appear in the new golden answer, while the other tokens are wrong and will be outdated tokens after editing. 
This proportion of the common tokens and outdated tokens suggests that not all tokens in outdated responses are wrong.
To further probe the effect of DISCO on the common tokens and the outdated tokens, we display the variation of the probability of both tokens, before and after applying DISCO.
As shown in Figure~\ref{fig: tokens_prob}, the common tokens receive the increasing probability, while the probability of outdated tokens decreases from 66.05\% to 4.52\%.
The variation indicates that DISCO could reserve the common tokens, and effectively reduce the probability of outdated tokens.
The effect of DISCO on both tokens demonstrates that DISCO could preserve the probability of correct tokens and reduce the probability of outdated tokens, and then improve the performance of Portability.

\begin{table}[t!]
\centering
\resizebox{0.435\textwidth}{!} {
\begin{tabular}{c|cc|ccc}
    \toprule
     \multirow{2}{*}{\textbf{ID}} & \multicolumn{2}{c|}{\textbf{Constraints}} & \multirow{2}{*}{\textbf{Portability}$\uparrow$} & \multirow{2}{*}{\textbf{OE}$\downarrow$} & \multirow{2}{*}{\textbf{TE}$\downarrow$} \\
    ~ & \textbf{Outdated} & \textbf{Target} & ~ & ~ & ~ \\
     \midrule
    1 & \checkmark & \checkmark & \textbf{63.87} / \textbf{33.46} & 5.78 & 6.09 \\
    2 & \scalebox{0.75}{\usym{2613}} & \checkmark & 62.86 / 32.21 & 6.59 & \textbf{6.01} \\ 
    3 & \checkmark & \scalebox{0.75}{\usym{2613}} & 62.97 / 31.92 & \textbf{5.65} & 8.09\\ 
    4 & \scalebox{0.75}{\usym{2613}} & \scalebox{0.75}{\usym{2613}} & 61.96 / 30.76 & 6.44 & 7.97 \\ 
    \bottomrule
\end{tabular}
}
\caption{
    Ablation experimental results of DISCO on the zsRE dataset with \texttt{LlaMa-2-7b}.
    Outdated and Target denotes the constraints to $\mathit{\Delta}(t)$ as stated in Equ.\ref{eq.7} and \ref{equ: limit}.
}
\label{tab: abl_results}
\end{table}
\subsection{Ablation} \label{sec: ablation}
\modi{
We explore the impact of both constraints, in Equ.\ref{equ: limit}, to $\mathit{\Delta}(t)$ in DISCO, and conduct experiments on the zsRE dataset with \texttt{LlaMa-2-7b}.
We list the results in Table~\ref{tab: abl_results} and display the impact of prepended paraphrased edited knowledge in Appendix.\ref{appendix: ablation}.
}

\modi{
As illustrated in Table.\ref{tab: abl_results}, we remove the constraint on tokens of outdated response (ID.2), the target of edited knowledge (ID.3), and remove both constraints (ID.4), while setting the whole DISCO as the baseline (ID.1).
%
When removing the constraint on outdated tokens, ID.2 suffers a more serious outdated issue and worse quality of response, compared to ID.1.
Similarly, ID.3 generates more tokens of the target of edited knowledge, after removing the constraints on the corresponding tokens.
Consequently, after removing both constraints, ID.4 generates the worst response and suffers the deterioration on both token types.
These performances suggest that both constraints are efficient in revising the probability of the corresponding type of tokens, while DISCO could improve the probability of golden answers.
}

\subsection{Efficiency}
Knowledge Editing should minimize the time required for conducting edits without compromising the model's performance.
We calculate the Time Cost for five methods on the \texttt{LlaMa-2-7b}, to compare the efficiency of DISCO.
Similar to \citet{yao2023editing}, we counter the average time-cost of 10 edits.

Table~\ref{tab: efficiency} illustrates the time required for different knowledge editing methods from providing the edited case to obtaining the edited model. 
We could observe that DISCO could quickly edit knowledge, much faster than other methods, and then yield remarkable performance (details of results as discussed in Section~\ref{sec: main_results}). 
On the other hand, ROME and MEMIT are time-consuming and necessitate a pre-computation of the covariance statistics for the Wikitext. 
Hence, considering the time aspect, DISCO is the optimal time-friendly knowledge editing method, with remarkable performance.

\begin{table}[t!]
\centering
\resizebox{0.38\textwidth}{!} {
\begin{tabular}{ccccc}
    \toprule
     \textbf{FT} & \textbf{ROME} & \textbf{MEMIT} & \textbf{IKE} & \textbf{DISCO} \\
    \midrule
    34.74s & 154.05s & 127.99s & 32.43s & \textbf{19.69}s \\
    \bottomrule
\end{tabular}
}
\caption{
    Average wall clock time for each edit method conducting 10 edits on LlaMa-2-7b, using singe A100 (40G).
}
\label{tab: efficiency}
\end{table}

\begin{table}[t!]
\centering
\resizebox{0.38\textwidth}{!} {
\begin{tabular}{c|cccc}
    \toprule
    \textbf{Method} & \textbf{Locality}$\uparrow$ & \textbf{Portability}$\uparrow$ & \textbf{OE}$\downarrow$ & \textbf{TE}$\downarrow$ \\
    \bottomrule
    \multicolumn{5}{c}{\texttt{LlaMa-2-7b}} \\
    \toprule
    IKE &  56.27 / 22.08 & 50.88 / 10.90 & \;\;8.23 & 17.64 \\
    DISCO & \textbf{62.64} / \textbf{30.95} & \textbf{63.87} / \textbf{33.46} & \;\;\textbf{5.78} & \;\;\textbf{6.09} \\
    \bottomrule
    \multicolumn{5}{c}{\texttt{LlaMa-2-7b}} \\
    \toprule
    IKE & 60.68 / 28.83 & 55.26 / 18.13 & \;\;7.14 & 15.45 \\ 
    DISCO & \textbf{61.31} / \textbf{29.80} & \textbf{66.85} / \textbf{37.61} & \;\;\textbf{5.26} & \;\;\textbf{5.04} \\
    \bottomrule
\end{tabular}
}
\caption{
    Experimental results on the zsRE dataset with \texttt{LlaMa-2-7b} and \texttt{LlaMa-2-13b}.
}
\label{tab: large_model}
\end{table}

\subsection{Model Scaling}
We apply two methods (\emph{i.e.}, IKE, and DISCO) to LlaMa-2-13b to evaluate the reasoning ability of the edited knowledge with model scaling.

As Table~\ref{tab: large_model} shows, when applying to LlaMa-2-13b, both methods perform better in Portability than in LlaMa-2-7b, with 3.29 and 1.50 F1 scores improvement, respectively.
Besides, both methods generate fewer outdated tokens and factual knowledge tokens. 
Unfortunately, DISCO performs worse in Locality on the larger LLM, suggesting that the edited knowledge has more impact on the probability distribution on the larger LLM, thus disturbing the LLM locating the editing scope.
Additionally, compared to the IKE with the LlaMa-2-13b, DISCO achieves better performance in the four properties on the LlaMa-2-7b.
The experimental results indicate that DISCO has a strong capability to enhance edited models on the reasoning problem w.r.t the edited knowledge and performs better with the larger model.

\section{Conclusion}
In this paper, we focus on the outdated issue that edited models suffer from generating outdated responses to the reasoning questions, which is ignored by previous approaches.
To mitigate this issue, we propose a simple yet effective method, out\textbf{D}ated \textbf{IS}sue aware de\textbf{CO}ding (\textbf{DISCO}), to encourage the edited model to utilize the edited knowledge to reason the correct answers for reasoning questions.
Specifically, DISCO captures and amplifies the modification in probability distribution between the original and edited models.
Experimental results demonstrate that DISCO could significantly mitigate the outdated issue, and effectively encourage the edited to reason the new correct answers for reasoning questions, without updating parameters.

\section*{Limitations}
In this paper, we investigate the outdated issue of edited models to the reasoning questions, and we propose a simple yet effective decoding strategy, \emph{i.e.}, DISCO, to prevent this issue and enhance the performance of edited models on reasoning questions.
In this paper, we mainly focus on the outdated issue in the one-hop reasoning questions, which is the prior part of multi-hop questions.
We leave these in future work to take further improvement.

\section*{Acknowledgements}
The research work described in this paper has been supported by the National Nature Science Foundation of China (No. 61976016, 62376019, 61976015), and the authors would like to thank the anonymous reviewers for their valuable comments and suggestions to improve this paper.

\bibliography{anthology,custom}
\bibliographystyle{acl_natbib}

\begin{table*}[t!]
\centering
\resizebox{0.95\textwidth}{!} {
\begin{tabular}{c|c|c|ccc|c|cc}
    \toprule
    \textbf{Backbone} & \makecell[c]{\textbf{Method}} & \textbf{Update?} & \textbf{Reliability}$\uparrow$  & \textbf{Generality}$\uparrow$ & \textbf{Locality}$\uparrow$ & \textbf{Portability}$\uparrow$ & \textbf{OE}$\downarrow$ & \textbf{TE}$\downarrow$ \\
    \bottomrule
    \multicolumn{9}{c}{\textit{zsRE}} \\
    \toprule
    %
    \multirow{6}*{\makecell[c]{\texttt{GPT-J-6b}}} & FT & \checkmark & 15.32 / 0.10\;\; & 14.73 / 0.00\;\; & 99.22 / 97.40 & 30.39 / 0.00 & 48.83 & \;\;\textbf{1.28} \\
    ~ & ROME & \checkmark & \textbf{99.56} / \textbf{99.13} & \textbf{92.66} / \textbf{88.33} & 80.15 / 60.56 & 31.70 / \textbf{2.70} & \textbf{15.62} & 21.80 \\
    ~ & MEMIT & \checkmark  & 98.93 / 98.07 & 80.60 / 69.24 & \textbf{99.26} / \textbf{97.69} & \textbf{33.01} / 1.35 & 30.35 & \;\;9.82 \\
    \cmidrule{2-9}
    ~ & IKE & \scalebox{0.75}{\usym{2613}} & \textbf{99.90} / \textbf{99.71} & \textbf{99.82} / \textbf{99.61} & 52.28 / 28.93 & 36.46 / 3.57 & 12.41 & 25.80 \\
    ~ & DISCO & \scalebox{0.75}{\usym{2613}} & 98.30 / 97.59 & 97.32 / 96.14 & \textbf{54.97} / \textbf{31.44} & \textbf{42.05} / \textbf{11.39} & \textbf{11.58} & \textbf{13.78} \\
    \midrule
    %
    %
    \multirow{6}*{\makecell[c]{\texttt{LlaMa-2-7b}}} & FT & \checkmark & 43.29 / 9.35\;\; & 36.80 / 4.34\;\; & 93.51 / 80.62 & 35.21 / 0.77 & 12.75 & 10.90 \\
    ~ & ROME & \checkmark & \textbf{75.28} / \textbf{64.32} & \textbf{70.39} / \textbf{55.35} & 97.22 / 90.74 & \textbf{37.28} / \textbf{2.86} & 14.51 & \;\;\textbf{7.67} \\
    ~ & MEMIT & \checkmark & 74.81 / 60.84 & 69.65 / 51.49 & \textbf{98.46} / \textbf{94.60} & 37.02 / 2.51 & \textbf{12.70} & \;\;9.72 \\
    \cmidrule{2-9}
    ~ & IKE & \scalebox{0.75}{\usym{2613}} & \textbf{99.84} / \textbf{99.71} & \textbf{99.63} / \textbf{99.32} & 56.27 / 22.08 & 50.88 / 10.90 & \;\;8.23 & 17.64 \\
    ~ & DISCO & \scalebox{0.75}{\usym{2613}} & 99.02 / 98.17 & 98.90 / 97.69 & \textbf{62.64} / \textbf{30.95} & \textbf{63.87} / \textbf{33.46} & \;\;\textbf{5.78} & \;\;\textbf{6.09} \\
    \bottomrule
    \multicolumn{9}{c}{\textit{CounterFact}} \\
    \toprule
    \multirow{6}*{\makecell[c]{\texttt{LlaMa-2-7b}}} & FT & \checkmark & 37.15 / 26.67 & 27.60 / 19.30 & 38.60 / 31.23 & 30.21 / 0.00 & 42.92 & \;\;1.65 \\
    ~ & ROME & \checkmark & 83.23 / \textbf{77.30} & 40.26 / 31.81 & 90.30 / 88.26 & \textbf{33.24} / \textcolor{gray}{0.00} & 36.53 & \textbf{0.80} \\
    ~ & MEMIT & \checkmark & \textbf{83.33} / \textbf{77.30} & \textbf{46.68} / \textbf{38.41} & \textbf{93.31} / \textbf{91.85} & 30.97 / \textcolor{gray}{0.00} & \textbf{33.04} & 1.92 \\
    \cmidrule{2-9}
    ~ & IKE & \scalebox{0.75}{\usym{2613}} & \textbf{99.44} / \textbf{99.32} & \textbf{82.00} / \textbf{79.44} & \textbf{25.86} / \textbf{17.75} & 38.05 / \textcolor{gray}{0.00} & 37.49 & 0.33 \\
    ~ & DISCO & \scalebox{0.75}{\usym{2613}} & 91.85 / 90.40 & 81.43 / 78.56 & 19.13 / 9.80\;\; & \textbf{39.41} / \textcolor{gray}{0.00} & \textbf{36.34} & \textbf{0.08} \\
    \bottomrule
\end{tabular}
}
\caption{
    Experimental results (F1/EM) on the \textit{zsRE} and \textit{CounterFact} datasets with \texttt{GPT-J-6b}, \texttt{LlaMa-2-7b}.
    All digital results denote the token-level score of the corresponding property.
    \textbf{Portability} is the core problem in this paper.
    \textbf{DISCO} denotes our decoding strategy.
    \textbf{Bold} denotes the best performance.
}
\label{tab: full_results}
\end{table*}

\begin{table*}[t!]
\centering
\resizebox{0.87\textwidth}{!} {
\begin{tabular}{c|cccc|cccc}
    \toprule
    \multirow{3}{*}{\textbf{Value}} & \textbf{Locality}$\uparrow$ & \textbf{Portability}$\uparrow$ & \textbf{OE}$\downarrow$ & \textbf{TE}$\downarrow$ & \textbf{Locality}$\uparrow$ & \textbf{Portability}$\uparrow$ & \textbf{OE}$\downarrow$ & \textbf{TE}$\downarrow$ \\
    \cmidrule{2-9}
    ~ & \multicolumn{4}{c|}{\texttt{GPT-J-6b}} & \multicolumn{4}{c}{\texttt{LlaMa-2-7b}} \\
    \midrule
     0.1 & \textbf{65.70} / \textbf{41.80} & 41.94 / 9.93\;\; & 15.61 & 14.61 & \textbf{70.07} / \textbf{37.99} & 63.65 / 31.92 & 6.63 & 7.42 \\
     0.3 & 61.82 / 37.13 & 42.03 / 10.22 & 14.06 & 14.72 & 67.92 / 35.58 & 63.99 / 32.79 & 6.24 & 7.12 \\
     0.5 & 59.50 / 35.00 & 42.05 / 10.51 & 13.10 & 14.67 & 66.26 / 34.23 & \textbf{64.11} / 33.27 & 6.02 & 6.74 \\
     \;\,1.0* & 54.97 / 31.44 & 42.05 / 11.28 & 11.39 & 14.38 & 62.72 / 31.05 & 63.87 / \textbf{33.46} & 5.78 & 6.09 \\
     1.5 & 51.16 / 28.64 & 42.14 / \textbf{12.15} & 10.54 & 14.21 & 59.70 / 28.54 & 63.84 / \textbf{33.65} & 5.29 & 5.78 \\
     2.0 & 48.42 / 26.90 & \textbf{42.26} / 12.05 & \;\;\textbf{9.93} & \textbf{13.81} & 57.64 / 27.58 & 63.63 / 33.08 & \textbf{5.04} & \textbf{5.32} \\
    \bottomrule
\end{tabular}
}
\caption{
    Experimental results of DISCO with different values of hyperparameter $\alpha$ on the zsRE dataset on the \texttt{GPT-J-6b} and \texttt{LlaMa-2-7b}.
    Digits in the \textbf{Value} column present the selection of the hyperparameter $\alpha$, and the other digital results denote the token-level score of the corresponding property.
    * denotes the default selection of hyperparameter $\alpha$.
    \textbf{Bold} denotes the best performance.
}
\label{tab: hyper_results}
\end{table*}

\begin{table}[t!]
\centering
\resizebox{0.365\textwidth}{!} {
\begin{tabular}{c|ccc}
    \toprule
     \textbf{Model} & \textbf{Portability}$\uparrow$ & \textbf{OE}$\downarrow$ & \textbf{TE}$\downarrow$ \\
     \midrule
    DISCO & \textbf{63.87} / \textbf{33.46} & \textbf{5.78} & 6.09 \\
    DISCO \textit{w.o.} prepended & 63.18 / 33.26 & 6.00 & 4.92 \\ 
    \bottomrule
\end{tabular}
}
\caption{
    Ablation Experimental on the impact of the prepended paraphrased edited knowledge to DISCO on the zsRE dataset with \texttt{LlaMa-2-7b}.
}
\label{tab: case_results}
\end{table}

\newpage
\appendix

\section{Full Comparison in zsRE} \label{sec: full}
We supply the experimental results of ROME and MEMIT in Table~\ref{tab: full_results}. 
As Table~\ref{tab: full_results} illustrated, we could observe that ROME and MEMIT perform well in Portability. 
With the benefit of the pre-computation of the covariance statistics, ROME and MEMIT could precisely update the parameters related to the edited knowledge, with more editing time-cost. 
As a result, ROME and MEMIT could retain the most performance when receiving the input out of the editing scope.

\section{Hyperparameter} \label{appendix: hyper}
To explore the impact of hyperparameter $\alpha$ in DISCO, we conduct experiments in the zsRE dataset on both backbones.
Since the F1 scores of Reliability and Generality are over 98, we pay less attention to both properties and mainly focus on the other metrics.

We adjust $\alpha$ to $(0.1, 0.3, 0.5, 1.0, 1.5, 2.0)$ to observe the corresponding performance of DISCO and report the results in Table~\ref{tab: hyper_results}.
As the results show, with the larger value of $\alpha$, the edited models generate the less outdated tokens, where ratios of Outdated issue downwards from 15.61 to 9.93 and 6.63 to 5.04 on both backbones, respectively.
Besides, the larger $\alpha$ hinders edited models from generating factual knowledge tokens, where ratios of \textit{TE} downwards from 14.61 to 13.81 and 7.42 to 5.32 on both backbones, respectively.
However, the larger $\alpha$ cannot yield better response quality, and the EM score is optimal when alpha is around 1.0. 
Unfortunately, the larger $\alpha$ gives rise to the aggravation of the performance in Locality, where the $\mathit{\Delta}$ in DISCO will lower the probability of the tokens predicted by the original model.
However, we could obtain the trade-off between Locality and Portability by setting $\alpha$ to 1.0.
With this setting, DISCO could yield competitive and stable performance with IKE in Locality, and outperform IKE in Portability.

\section{Ablation for Prepended Paraphrased edited knowledge} \label{appendix: ablation}
We supply the exploration of the impact of the prepended paraphrased edited knowledge to DISCO on the zsRE dataset with \texttt{LlaMa-2-7b}.
As illustrated in Table \ref{tab: case_results}, without the paraphrased edited knowledge, DISCO suffers a loss in terms of portability and outdated issue problems, compared to DISCO.
The worse performance of DISCO without the paraphrased edited knowledge demonstrates that the prepend paraphrased edited knowledge is beneficial for the edited model and enhances its performance on reasoning questions w.r.t. edited knowledge.

\end{document}